\documentclass[sigconf, authorversion, screen, nonacm]{acmart}
\AtBeginDocument{%
  \providecommand\BibTeX{{%
    \normalfont B\kern-0.5em{\scshape i\kern-0.25em b}\kern-0.8em\TeX}}}

\setcopyright{acmcopyright}
\copyrightyear{2018}
\acmYear{2018}
\acmDOI{XXXXXXX.XXXXXXX}

\acmConference[Conference acronym 'XX]{Make sure to enter the correct
  conference title from your rights confirmation emai}{June 03--05,
  2018}{Woodstock, NY}
\acmPrice{15.00}
\acmISBN{978-1-4503-XXXX-X/18/06}

\usepackage{multirow}
\usepackage{subcaption}
\newcommand{\Lagr}{\mathcal{L}}
\usepackage{amsthm}
\newtheorem{definition}{Definition}
\newtheorem{problem}{Problem}

\makeatletter
\newcommand\footnoteref[1]{\protected@xdef\@thefnmark{\ref{#1}}\@footnotemark}
\makeatother

\begin{document}

\title[Improving next location prediction]{How do you go where? Improving next location prediction by learning travel mode information using transformers}

\author{Ye Hong}
\email{hongy@ethz.ch}
\orcid{0000-0002-8996-3748}
\affiliation{%
  \institution{Institute of Cartography and Geoinformation, ETH Zurich}
  \city{Zurich}
  \country{Switzerland}
}

\author{Henry Martin}
\email{martinhe@ethz.ch}
\orcid{0000-0002-0456-8539}
\affiliation{%
  \institution{Institute of Cartography and Geoinformation, ETH Zurich}
  \city{Zurich}
  \country{Switzerland}
}
\affiliation{%
  \institution{Institute of Advanced Research in Artificial Intelligence (IARAI)}
  \city{Vienna}
  \country{Austria}
}

\author{Martin Raubal}
\email{mraubal@ethz.ch}
\orcid{0000-0001-5951-6835}
\affiliation{%
  \institution{Institute of Cartography and Geoinformation, ETH Zurich}
  \city{Zurich}
  \country{Switzerland}
}


\begin{abstract}
Predicting the next visited location of an individual is a key problem in human mobility analysis, as it is required for the personalization and optimization of sustainable transport options.
Here, we propose a transformer decoder-based neural network to predict the next location an individual will visit based on historical locations, time, and travel modes, which are behaviour dimensions often overlooked in previous work.
In particular, the prediction of the next travel mode is designed as an auxiliary task to help guide the network's learning.
For evaluation, we apply this approach to two large-scale and long-term GPS tracking datasets involving more than 600 individuals.
Our experiments show that the proposed method significantly outperforms other state-of-the-art next location prediction methods by a large margin ($8.05\%$ and $5.60\%$ relative increase in F1-score for the two datasets, respectively).
We conduct an extensive ablation study that quantifies the influence of considering temporal features, travel mode information, and the auxiliary task on the prediction results.
Moreover, we experimentally determine the performance upper bound when including the next mode prediction in our model.   
Finally, our analysis indicates that the performance of location prediction varies significantly with the chosen next travel mode by the individual.
These results show potential for a more systematic consideration of additional dimensions of travel behaviour in human mobility prediction tasks. 
The source code of our model and experiments is available at \url{https://github.com/mie-lab/location-mode-prediction}.

\end{abstract}




\maketitle

\section{Introduction}

The rapid urbanization process in the last decades has caused a constant increase in individual travel, imposing significant challenges in achieving sustainable cities. To meet the sustainable development goals of the United Nations~\citep{griggs2013sustainable}, mobility behaviour change and new mobility concepts to promote these changes will play indispensable roles~\citep{martin2021using}.
These mobility concepts, such as mobility as a service (MaaS)~\citep{reck_mode_2022}, smart charging~\citep{xu_planning_2018} and ride-sharing~\citep{huang_multimodal_2019}, all rely on the capability to proactively provide personalized services that are tailored to the travel context and individuals’ characteristics~\citep{ma_individual_2022}. Individual mobility prediction to know when and where travel will occur is a crucial technique driving the development and application of these new concepts and, therefore, a key technology for sustainable transportation. 

The prediction of where an individual will go given her historical mobility information is central to individual mobility prediction. The problem is also known as the next location prediction and has attracted much attention over the last decade. Researchers are increasingly interested in tackling this problem using learning-based methods thanks to the booming of deep learning (DL) models~\citep{luca_survey_2021}. 
As it can be formulated as a sequence prediction problem, similar to the tasks encountered in natural language processing and audio processing, models that have shown success in these two fields are often directly applied. 
In particular, the transformer model~\citep{Vaswani_2017} that utilizes a multi-head self-attention mechanism has revolutionized various sequence modelling tasks due to its powerful and efficient network structure. Transformer models are also starting to gain attention in predicting individual mobility, as they tackle some challenges in mobility prediction by design:
(1) Multiple periodicities co-exist in the location visitation patterns of individuals~\citep{FengLZSMGJ18} (e.g., daily, weekly). These periodicities vary considerably across individuals. The multi-head self-attention module allows the network to focus on multiple steps in the input sequence, effectively capturing these periodicities.
(2) The long-term dependency of mobility behaviour. Studies have shown that the current mobility depends on behaviours conducted days or weeks before~\citep{sun_understanding_2013, cherchi_modelling_2017}, which requires the prediction model to capture long-term dependencies. The design of the transformer model enables efficient learning of these dependencies.
However, human mobility also exhibits unique characteristics, such as complex spatio-temporal dependencies~\citep{FengLZSMGJ18, li_hierarchical_2020} and the inherent stochasticity of location visits~\citep{song2010limits}, which hinder the performance when applying sequence learning models on raw location visit sequences.
Therefore, learning to predict the next location directly from historical location visits is challenging. An accurate prediction model should consider context information that influences individuals' choice of locations.

Results from travel behaviour studies that aim to understand individuals' activity location choices could guide the consideration of context information. Empirical evidence suggests that the selection of activity locations is highly correlated to other aspects of individual travel behaviour, such as the availability of travel modes~\citep{neutens_spacetime_2007} and the day of the week~\citep{dharmowijoyo_day--day_2016}.
However, the comprehensive information regarding individuals' travel behaviour is not fully utilized in location prediction problems. To date, it is still unclear (1) how strong the influence of these long-term factors is on choosing the immediate next location and (2) whether the DL network can benefit from this knowledge and learn the complex dependency patterns directly from data.  

To close this research gap and answer the above questions, we propose a transformer-based model that utilizes historical travel behaviour to predict individuals' next location. More precisely, the model aims to learn mobility transition patterns from historical location, temporal and travel mode sequence information. Inspired by travel behaviour studies, we encourage the model to also predict the next travel mode the individual will choose. We anticipate that this ancillary task will help the prediction of the next location. Through experiments on two real-world GPS datasets, we demonstrate the effectiveness of our model design and quantify the dependency of location prediction performance on travel mode. Our results show that careful consideration of individual travel behaviour significantly benefits human mobility prediction.
In short, our contributions are summarized as follows:
\begin{itemize}
  \item We propose a transformer decoder-based neural network that utilizes location, travel mode and time-related information for the next location prediction task. The proposed model achieves state-of-the-art performance.
  \item We show that jointly learning the next location and next mode improves the prediction performance for both tasks.
  \item We conduct extensive experiments on two real-world GPS tracking datasets and conclude that considering additional aspects of travel behaviour significantly increases the performance of next location prediction.
\end{itemize}

The rest of this paper is organized as follows. We first systematically review related work in Section 2. In Section 3, we formulate the next location prediction problem. Next, we introduce details of the network architecture in Section 4. We apply our model to two real-world GPS datasets and analyze its performance in Section 5. Finally, we summarize the main findings and conclude the paper in Section 6.

\section{Related work}

\subsection{Next location prediction}
The next location prediction problem has found application in many different fields, such as recommendation systems~\citep{xue2021mobtcast}, sensor networks~\citep{pirozmand_human_2014}, and mobility behaviour analysis~\citep{xu_understanding_2022, WangW0ZHF21}. 
The exact definition of the problem varies across studies due to different objectives and employed datasets. For example, location-based social network (LBSN) applications focus on predicting the next check-in point-of-interest (POI)~\citep{xue2021mobtcast, Wang_2022}. In contrast, mobility behaviour studies aim to understand the next location for a user to conduct an activity~\citep{solomon_analyzing_2021}. Here we focus on the methods proposed for mobility applications. 

The last decade has witnessed the expansion of studies focusing on next location prediction. Markov Chain and its variants are probably the most often employed methods for the task~\citep{luca_survey_2021}. These models regard locations as states and construct a transition matrix that encodes the transition probability between states for each individual. \citet{AshbrookS02} and \citet{Gambs_2012} both proposed identifying significant locations from GPS data and building a Markov model to predict location transitions. 
Later Markov model variants that consider collective movements~\citep{ChenLY14} and incorporate location importance~\citep{huang_mining_2017} further increased the prediction performance. However, Markov-based models struggle to represent the complex sequential patterns in human mobility because of their inherent assumption that the current state only depends on the states of previously limited time steps~\citep{li_hierarchical_2020}. 

Recent advances in DL have also promoted their application in location prediction. As a widely adopted sequence modelling method, recurrent neural network (RNN)-based models, such as Long Short-Term Memory (LSTM)~\citep{solomon_analyzing_2021} and spatial-temporal (ST)-RNN~\citep{LiuWWT16}, were reported to outperform Markov models by a large margin in the task. Still, vanilla RNN models tend to underweight long-term dependencies when the input sequence length increases. Therefore, studies employed the attention module to capture both short-term and long-term dependencies dynamically~\citep{FengLZSMGJ18, li_hierarchical_2020}.
Moreover, the transformer model that builds on top of the multi-head self-attention mechanism~\citep{Vaswani_2017} have started to gain interest in the field. In particular, \citet{xue2021mobtcast} proposed MobTcast for considering various contexts with a transformer-based structure and achieved state-of-the-art POI prediction results for LBSN data. 
Although having great potential in learning the complex spatio-temporal dependencies, limited studies have applied transformer for the location prediction problem. 

\subsection{Factors affecting activity location choice}

Understanding the factors affecting activity location choice is beneficial for predicting individuals' mobility, as they can be regarded as prior knowledge and potentially guide the learning of DL models. 
In the travel behaviour field, the choice of locations is regarded as an integral part of individuals' activity-travel behaviour and has been studied within the activity-based framework~\citep{schonfelder_urban_2016}. 
Studies that focus on analysing travel behaviour over time suggest that both stability and variability are found in individuals' activity location choices. For example, \citet{dharmowijoyo_analysing_2017} showed that the variability of location visits is much larger between weekend-weekday pairs than between weekday-weekday and weekend-weekend pairs.
Empirical studies also demonstrate the correlation of different aspects of individual travel behaviour. For example, \citet{susilo_repetitions_2014} reported high repetition in location-mode combinations, suggesting that individuals use the same travel mode to reach their locations. Similar conclusions were reported by~\citet{Hong_2022}, where they found that only a subset of all location-mode combinations is essential for describing the mobility behaviour. From this perspective, aspects of travel behaviour can be considered constraints for individuals' choice of activity locations.

A similar problem as the next location prediction is the formulation of an individual's location choice set, which is a crucial component in microscopic traffic simulation models~\citep{leite_mariante_modeling_2018}. Instead of predicting the exact next location, the problem aims at generating a set containing all possible locations. Based on time geography theory, potential path areas analysis has been applied to tackle the problem, suggesting that the choice set is constrained by the travel time~\citep{scott_modeling_2012}, time of day~\citep{yoon_feasibility_2012} and the available travel mode~\citep{neutens_spacetime_2007}. However, this empirical knowledge is not fully utilized in models for location predictions.

Building on the previous studies, we utilize a transformer-based model for learning spatio-temporal dependencies in individual location visits. In addition, we consider other aspects of travel behaviour that constrain an individual's activity location choices in the learning process.

\section{Problem formulation}
We formulate the next location prediction problem with the following definitions:

\begin{sloppypar}
    \begin{definition}[Trajectory]
        Let $u^{(i)}$ be a user in a set of users $\mathcal{U} = \left \{ u^{(1)}, u^{(2)}, ..., u^{(\left |\mathcal{U}  \right |)} \right \}$, a trajectory $T^{(i)} = [q_1, q_2, ..., q_{n_{u^{(i)}}}]$ is a time-ordered sequence composed of $n_{u^{(i)}}$ track points visited by $u^{(i)}$. A track point can be represented as a tuple of $q = \langle p, t\rangle $, where $t$ records the time when the user visits, and $p = \langle x, y\rangle $ represents spatial coordinates in a given reference system, e.g., latitude and longitude. 
    \end{definition}
\end{sloppypar}

\begin{definition}[Location sequence]
    Location is defined when a user remains within a certain geographical radius for a defined time, and is a sub-sequence of the user's trajectory. A location sequence $S^{(i)} = [L_1, L_2, ..., L_{w_{u^{(i)}}}]$ is a time-ordered sequence composed of $w_{u^{(i)}}$ locations visited by $u^{(i)}$. Each location $L = \langle l, t, e \rangle $ is described by the arrival time $t$, the (main) travel mode $e$ used to reach that location, and the identifier $l \in \mathcal{O}$, where $\mathcal{O}$ is the set containing all known locations.
\end{definition}


\begin{problem}[Next location prediction]
    Consider the historical location sequence $S^{(i)}_{hist} = [L_{n-m+1}, L_{n-m+2}, ..., L_{n}]$ that $u^{(i)}$ visited, where $n$ is the current time step and $m$ is the number of considered previous locations, the goal is to predict the location identifier the same user $u^{(i)}$ will visit in the next time step, i.e., $l_{n+1} \in \mathcal{O}$. 
\end{problem}

In this study, we construct the historical location sequence using locations visited in the previous 7 days. Thus, the number of considered previous locations $m$ depends on the user $u^{(i)}$ and the current time step $n$, making the next location prediction a sequence prediction problem with variable sequence length.

\section{Methods}
\begin{figure*}[!htb]
  \centering
  \includegraphics[width=\linewidth]{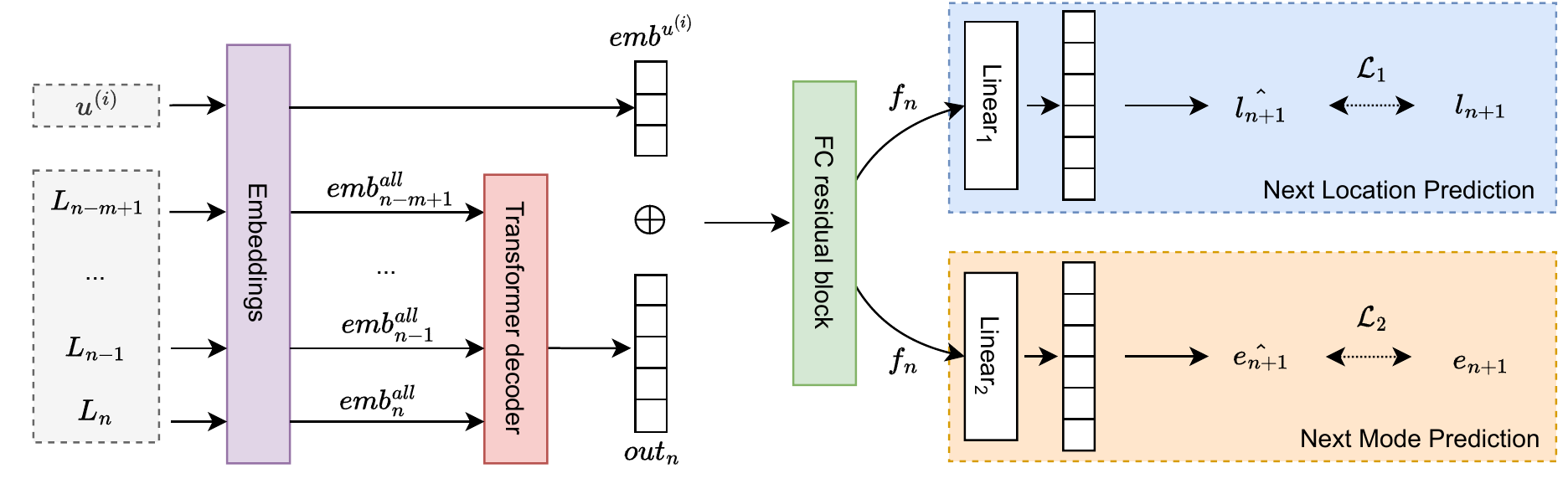}
  \caption{Structure of the proposed next location prediction model. The model learns to represent input features into embedding vectors, which are fed into a transformer decoder neural network. The network's output is concatenated with user embedding and processed through a fully connected residual block for predicting the next location. We introduce the next mode prediction as an auxiliary task to guide the learning process.}
  \label{fig:overview}
\end{figure*}

Figure~\ref{fig:overview} presents the overview of the proposed model. Specifically, the model considers historical location, time and travel mode patterns as input and learns from the auxiliary task of predicting the next travel mode to improve the prediction of the next location. The model consists of three major components: (1) feature embedding (Section~\ref{sec:emb}); (2) transformer decoder and fully connected (FC) layers (Section~\ref{sec:trans}); and (3) loss design and prediction (Section~\ref{sec:Loss}). The detailed description of each component is presented in the following subsections.

\subsection{Embedding learning}
\label{sec:emb}

\begin{figure}[htbp!]
  \centering
  \includegraphics[width=\linewidth]{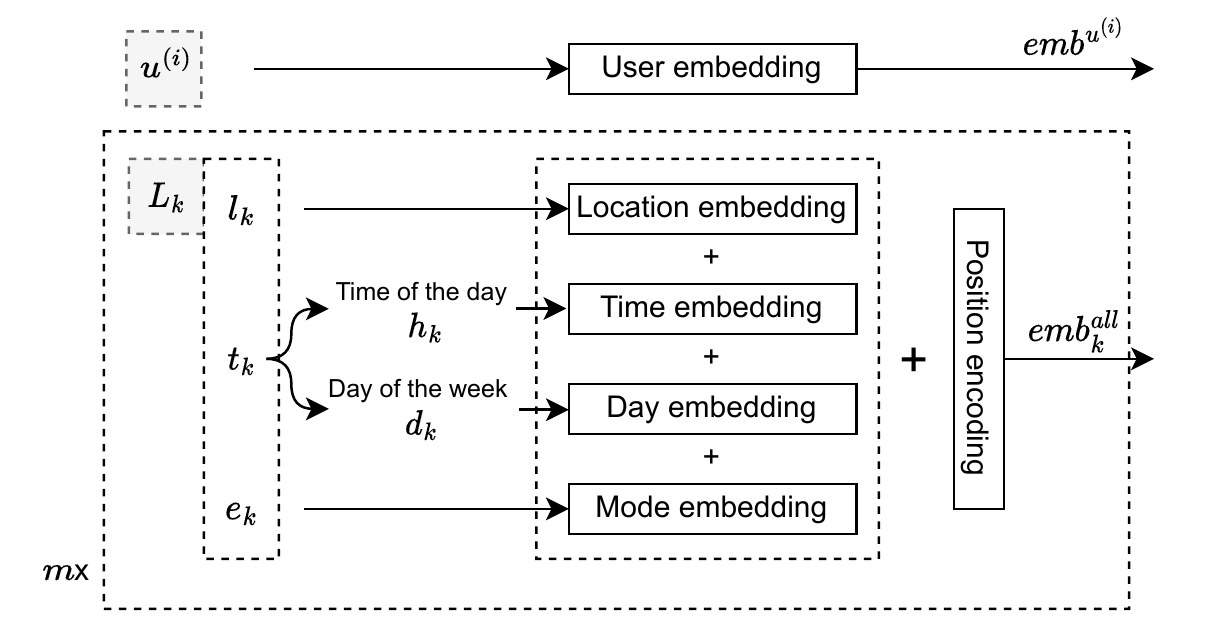}
  \caption{Pipeline of generating the embedding vectors.}
  \label{fig:embed}
\end{figure}

The representation and modelling of historical information is a vital step for accurately predicting the next location. Most existing prediction models utilize the location visitation sequence to understand the complex spatial-temporal dependencies. We additionally consider temporal and travel mode information that can provide context for visitation patterns. Moreover, user information is beneficial for the network to identify location sequences travelled by different users and learn user-specific movement patterns.
Therefore, we regard the location $l_{k}$, the start time $t_{k}$, and the travel mode $e_{k}$ at any time step $k$ as well as the user $u^{(i)}$ (represented as a unique identifier) of the current sequence as the input features (see Figure~\ref{fig:embed}). Specifically, we extract the time of the day $h_{k}$ (grouped into 15-minute bins) and the day of the week $d_{k}$ from the start time $t_{k}$, aiming to separate different levels of periodicity in location visits. 

\begin{sloppypar}
We now introduce the embedding method to transform each feature from the categorical type to a finite-dimensional real-valued vector. This is a more efficient and effective way to represent the correlation between different categorical values compared to the commonly used one-hot encoding method~\citep{li_hierarchical_2020, FengYXYWL20}. The embedding layers are parameter matrices that provide mappings between the original variable and the real-valued vector, whose parameters are jointly optimized with the entire network. The feature construction process can be formulated as follows: 
\begin{align}
    emb^{l}_{k} &= l_{k}\mathbf{W_{l}} \\ 
    emb^{h}_{k} &= h_{k}\mathbf{W_{h}} \\
    emb^{d}_{k} &= d_{k}\mathbf{W_{d}} \\
    emb^{e}_{k} &= e_{k}\mathbf{W_{e}} \\
    emb^{u^{(i)}} &= u^{(i)}\mathbf{W_{u}}
\end{align}
where $emb^{l}_{k},\:emb^{h}_{k},\:emb^{d}_{k},\:emb^{e}_{k} \in \mathbb{R}^{d_{base}}$ and $emb^{u^{(i)}}\in \mathbb{R}^{d_{user}}$ are the respective embedding vectors, and $l_{k},\:h_{k},\:d_{k},\:e_{k}$ and $u^{(i)}$ are the respective one-hot encoded original categorical features. $\mathbf{W}$ terms stand for weight matrices that are optimized during training. See Figure~\ref{fig:embed} for an illustration of the embedding process. Finally, the total embedding vector $emb^{all}_{k}$ for each time step $k$ is obtained by adding all sequence features together with a position encoding $PE$:
\end{sloppypar}
\begin{equation}
emb^{all}_{k} = emb^{l}_{k} + emb^{h}_{k} + emb^{d}_{k} + emb^{e}_{k} + PE
\end{equation}

We use the original positional encoding proposed by~\citet{Vaswani_2017} that utilizes sine and cosine functions to encode sequence information in the embedding. Positional encoding is essential for the training as the self-attention module does not implicitly assume sequential order~\citep{Vaswani_2017}. Note that we use addition instead of a concatenation operation to combine embedding vectors~\citep{zhou2021informer}. This process allows the model to flexibly balance the importance of each feature, as compared to manually assigning and tuning the sizes for different embedding vectors. 
After the embedding process, we obtain a single user embedding $emb^{u^{(i)}}$ and a sequence of embedding vectors $emb^{all}_{k}$ representing features at each time step $k$, which can be further processed with the transformer-based network.

\subsection{Transformer and fully connected layers}
\label{sec:trans}

\begin{figure}[htbp!]
  \centering
  \includegraphics[width=0.9\linewidth]{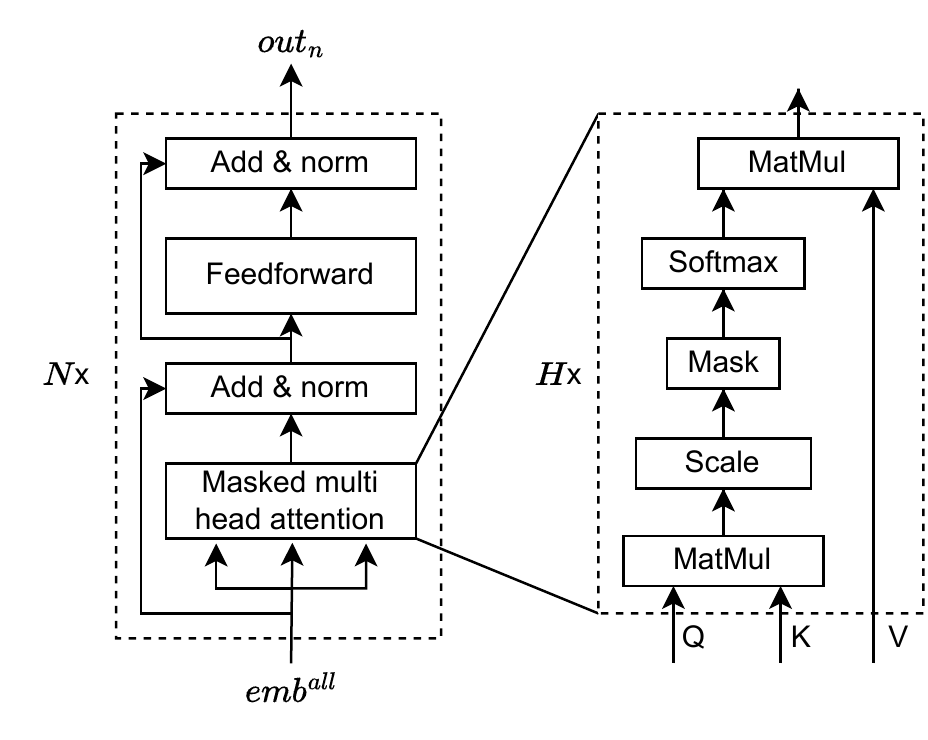}
  \caption{Structure of the transformer decoder neural network and the masked multi-head attention.}
  \label{fig:transformer}
\end{figure}

An accurate next location prediction model should be able to extract regularities and capture the multilevel periodicity from the complex spatio-temporal historical sequences~\citep{FengLZSMGJ18}. We utilize a transformer-based network to learn location transition patterns from historical location, temporal and travel mode information encoded in the final embedding vector $emb^{all}_{k}$. We adopt an architecture similar to the Generative Pre-trained Transformer (GPT) model, originally designed for language modelling~\citep{radford2018improving}, where only the transformer decoder part is used. The decoder consists of a stack of $N$ identical blocks (see Figure~\ref{fig:transformer}), each with two layers. The first is the masked multi-head attention, and the second is the fully-connected feedforward network with two linear layers and a ReLU activation function. Residual connections and layer normalization components are added to each layer to facilitate learning~\citep{He_16}. The input and output dimension of each block is designed to be the same as the embedding vector, i.e., $d_{model} = d_{base}$.

The core of the transformer structure is the multi-head self-attention mechanism. The attention function can be understood as obtaining an output value based on a query and a set of key-value pairs, all of which are vectors of size $d_k$. As described by~\citet{Vaswani_2017}, transformer uses the scaled dot-product attention and implements the calculation efficiently by packing the set of query, key and value values into matrices Q, K and V. This process can be formulated as follows: 
\begin{equation}
    \text{Attention(}Q, K, V\text{)} = \text{softmax(}\frac{QK^T}{\sqrt{d_k}})V
\end{equation}

\begin{sloppypar}
    Then, multi-head attention is constructed by concatenating the results of $H$ attention functions (see Figure~\ref{fig:transformer}):
    \begin{align}
        \text{MultiHead(}Q, K, V \text{)} &= (head_1 \oplus ... \oplus head_H)\boldsymbol{W}^O \\
        \text{where}\, head_i &= \text{Attention(}Q\boldsymbol{W}^Q_i, K\boldsymbol{W}^K_i, V\boldsymbol{W}^V_i ) 
    \end{align}
    where $\oplus$ represents the concatenation operation and $\boldsymbol{W}$ terms are parameter matrices learned by the network. In each masked multi-head attention, key, value and query matrices are identical, all corresponding to the output of the previous block. In the first block, they are set as the embedding matrix $emb^{all}$, obtained through stacking all embedding vectors according to their sequence. Note that we additionally include the forward-masking operation to prevent the attention function from accessing information from ``future'' time steps (see Figure~\ref{fig:transformer}); that is, the entry at time step $i$ can only focus on the information preceding (and including) $i$. 
    The self-attention mechanism enables the model to access information from every step in the historical sequence and evaluate its importance by learning parameter matrices. This ensures that long-term dependencies in the historical pattern can be extracted. Additionally, thanks to the multi-head design, the model retains multiple sets of parameter matrices that focus on various places in the historical sequence, efficiently capturing the multi-level periodic nature of human mobility.
\end{sloppypar}

\begin{figure}[htbp!]
  \centering
  \includegraphics[width=\linewidth]{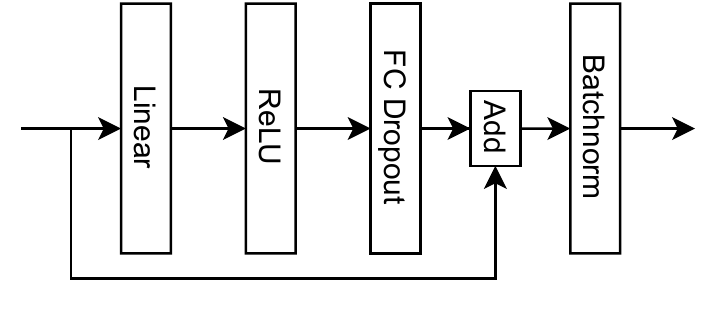}
  \caption{Structure of the fully-connected residual block.}
  \label{fig:fc}
\end{figure}

Then, the output vector of the transformer-decoder model $out_{n}$ is concatenated with the user embedding $emb^{u^{(i)}}$, and fed into a fully-connected residual block: 
\begin{equation}
    \label{equation:fc}
    f_n = FC(out_{n} \oplus emb^{u^{(i)}})
\end{equation}
where $FC(\cdot )$ represents the operation by the fully-connected residual block, whose structure is shown in Figure~\ref{fig:fc}. It consists of a single linear feedforward layer and a ReLU activation function, followed by a dropout layer, residual connection and a batch normalization layer. This block learns the dependencies of the transformer decoder output and the user embedding, helping the model extract user-specific movement patterns. Finally, the aggregated vector representation $f_n$ encapsulates location, temporal, travel mode and user information of the entire historical sequence, and can be further used to predict the next location.

\subsection{Loss and training}
\label{sec:Loss}

\begin{sloppypar}
The aggregated vector representation $f_n$ is then fed into two branches: the next location prediction branch and the next mode prediction branch. For the first task, we calculate the probability of each location through a linear projection and a softmax function: 
\begin{equation}
    P(\hat{l_{n+1}}) = \text{Softmax(Linear}_1(f_n))
\end{equation}
where $P(\hat{l_{n+1}})$ of size $\left | \mathcal{O}  \right |$ contains the probability of all locations to be visited at the next time step. Therefore, the most likely visited location $\hat{l_{n+1}}$ at time step $n+1$ is the location with the highest probability in $P(\hat{l_{n+1}})$. 
\end{sloppypar}

Given the ground truth next location $l_{n+1}$ from the training dataset, the task can be regarded as a multi-class classification problem. We obtain our location prediction loss $\Lagr_1$ using the multi-class cross entropy: 
\begin{equation}
    \Lagr_1 = -\sum_{i=1}^{\left |\mathcal{O}  \right |}P(l_{n+1})^{(i)}\log(P(\hat{l_{n+1}})^{(i)})
\end{equation}
where $P(\hat{l_{n+1}})^{(i)}$ represents the predicted probability of visiting the $i$-th location and $P(l_{n+1})^{(i)}$ is the one-hot represented ground truth, i.e., $P(l_{n+1})^{(i)}=1$ if the actual visited next location is the $i$-th location, and $P(l_{n+1})^{(i)}=0$ otherwise.

In addition to predicting the next location, we enforce the network to predict the next travel mode and optimize the network's parameters based on the joint prediction loss. This design is motivated by the assumption that learning to correctly predict the next travel mode can improve the prediction of the next location. Analogous to the next location probability, the probability of choosing each travel mode $P(\hat{e_{n+1}})$ can be obtained as follows:
\begin{equation}
    P(\hat{e_{n+1}}) = \text{Softmax(Linear}_2(f_n))
\end{equation}

Therefore, we introduce the mode prediction loss $\Lagr_2$ using the multi-class cross entropy:
\begin{equation}
    \Lagr_2 = -\sum_{i=1}^{\left |\mathcal{M}  \right |}P(e_{n+1})^{(i)}\log(P(\hat{e_{n+1}})^{(i)})
\end{equation}
where $\left |\mathcal{M}  \right |$ is the total number of travel mode categories, $P(\hat{e_{n+1}})^{(i)}$ represents the probability of using the $i$-th mode at time step $n+1$ and $P(e_{n+1})^{(i)}$ is the one-hot represented true next mode.

The final loss is the combination of the two losses. Given training samples in mini-batches of size $B$, the network's parameters are optimized by minimizing the following loss:
\begin{equation}
    \label{equation:final_loss}
    \Lagr = \sum_{b=1}^{B}(\Lagr_1(b) + \theta \Lagr_2(b))
\end{equation}
where $\Lagr_1(b)$ and $\Lagr_2(b)$ correspond to the location prediction loss and the mode prediction loss of the $b$-th training sample in the mini-batch, respectively, and $\theta$ balances the relative weights of the two losses. 



\section{Experiments}

\subsection{Datasets}

We test the proposed method using longitudinal tracking datasets from two tracking studies performed in Switzerland. 
    
\textbf{The Green Class (GC) study}~\citep{martin_begleitstudie_2019}. The study contains 139 participants based in Switzerland who got access to a comprehensive mobility package consisting of a general public transport pass valid in Switzerland, as well as access to several car- and bike-sharing programs and a battery electric vehicle for their personal use. 
All participants were asked to record their movements using a tracking app on their phones between November 2016 and January 2018. The app pre-processed the movement of participants into staypoints (stationary behaviour) and triplegs (continuous movement without changing travel mode). The study participants provided high-level activity labels for staypoints (\textit{home, work, errand, leisure, wait, and unknown}) and mode labels for triplegs (\textit{car, e-car, train, bus, tram, bicycle, e-bike, walk, airplane, boat, coach}).

\textbf{The yumuv study}~\citep{reck_mode_2022}. The study involves 498 participants based in and around the city of Z\"urich, Switzerland. Participants were tracked over three months between July and September 2020,  when COVID-19 case numbers, non-medical interventions and impact on people's daily lives were low in Switzerland. Participants got access to a mobility-as-a-service app called yumuv that provided simplified access to several micro-mobility modes. All participants were tracked via an app on their phones. Similar to the GC study, the app pre-processed movements into staypoints and triplegs that participants labelled with activity labels and travel modes.

We pre-process the raw movement data from the GPS tracking studies into corresponding analysis units for the next location prediction task using Python and the open-source \textit{Trackintel} framework~\citep{trackintel_arxiv}. We only consider users tracked for more than 300 days in GC and more than 30 days in Yumuv to ensure high temporal tracking coverage. 93 users in GC and 422 users in Yumuv remain. Based on the activity label, we regard a stay point as an activity if its duration is longer than 25 minutes or if it was labelled with a non-trivial purpose (any available purpose except for \textit{wait} or \textit{unknown}). Then, the activity stay points are spatially aggregated into locations to account for visits to the same place at different times. We utilized the function provided in \textit{Trackintel} with parameters \(\epsilon = 20\) and \(num\_samples = 2\) to generate \textit{dataset} locations~\citep{hong2021clustering}. Locations are also attached with the arrival travel mode. Since travel to a location may involve multiple stages with various travel modes, we determine the main travel mode as the mode with the longest distance~\citep{axhausen2007definition}. We further group the main travel mode into seven groups: \textit{walk, bicycle, train, tram, bus, car} and \textit{other} (including \textit{ski}, \textit{airplane} and \textit{coach}). Table~\ref{tab:number} shows the number of locations in both datasets, categorized based on the main travel mode.

\begin{table}[htbp!]
\caption{Main travel mode frequency of locations.}
\label{tab:number}

    \begin{tabular}{@{}ccc@{}}
    \toprule
    \textbf{Mode category}               & \textbf{GC}              & \textbf{Yumuv}                 \\ \midrule
    Walk                        & 55,000                    & 80,617                 \\
    Bicycle                     & 4,926                     & 24,745                 \\
    Train                       & 25,495                    & 20,260                 \\
    Tram                        & 2,182                     & 10,943                 \\
    Bus                         & 3,465                     & 10,962                 \\
    Car                         & 88,664                    & 55,289                 \\
    Other                       & 1,751                     & 476                   \\
    \textbf{Total}                       & \textbf{181,483}                   & \textbf{203,292}                \\ \bottomrule
    \end{tabular}
\end{table}

We split the location visitation sequence of each user into train, validation and test sets with the ratio of 6:2:2 based on time, such that location records that occurred in the first 60\% of days are considered as train and the last 20\% of days as test. The hyper-parameters are tuned on the validation set, and the test set is only used for reporting the prediction performance. We train a single population-level model using records from the training set of all users. We do not pre-filter the locations visited less often, which is a pre-processing step usually found in the location prediction models using GPS traces (e.g., \citep{solomon_analyzing_2021}). Although preserving the original location transition patterns, this setting makes the problem a more challenging task, as a large proportion of mobility consists of exploring new locations that are difficult to predict~\citep{cuttone_understanding_2018}.

\subsection{Setup}

The hyper-parameters for the network are reported in Table~\ref{tab:hyper}. We chose the best performing set of hyper-parameters using grid search: the number of layers $L$ from $\{2, 4, 6\}$, the number of heads $H$ from $\{4, 8\}$, the size of the embedding $d_{base}$ from $\{32, 64, 128, 256\}$ and the dropout of the FC layer from $\{0.1, 0.2, 0.5\}$. The $\theta$ in the final loss (Eq.~(\ref{equation:final_loss})) is set to 1. Adam optimizer with an initial learning rate of $1e^{-3}$ and L2 penalty of $1e^{-6}$ is used to optimize the model's parameters. We implement learning rate warm-up and decay as reported by~\citet{Vaswani_2017}. To alleviate the model over-fit on the training dataset, we use an early stopping strategy to pause the learning if the validation loss stops decreasing for 3 epochs. We then drop the learning rate by 0.1 and continue the training from the model with the lowest validation loss. This process is repeated 3 times.

\begin{table}[htbp!]
\tabcolsep=0.11cm
\centering
\caption{Experiment settings and network parameters.}
\label{tab:hyper}
    \begin{tabular}{@{}cccc@{}}
    \toprule
    \textbf{Setting} &\textbf{ Value}         & \textbf{Network parameter}      & \textbf{Value} \\ \midrule
    Learning rate (lr)  & $1e^{-3}$          & Emb. $d_{base}$         & 64     \\
    L2 penalty          & $1e^{-6}$          & User emb. $d_{user}$         & 16     \\
    lr warm up          & 2 epochs      & \# layers $N$              & 4      \\
    lr decay             & 0.02          & \# heads   $H$              & 8      \\
    Early stop patiance & 3 epochs      & Feedforward   & 256    \\
    Early stop lr drop  & 0.1           & FC dropout & 0.1    \\ \bottomrule
    \end{tabular}
\end{table}


\begin{table*}[ht]
\centering
\caption{Performance evaluation results for next location prediction. The mean and the standard deviation (in parentheses) are reported. Numbers marked in \textbf{bold} and \underline{underline} represent the best and the second-best performing method respectively. }
\label{tab:performance}
\begin{tabular}{@{}ccccccc@{}}
\toprule
Dataset & Method    & F1 & Acc@1 & Acc@5 & Acc@10 & MRR       \\ \midrule
\multirow{6}{*}{GC} & 1-MMC~\citep{Gambs_2012}             & 25.56 & 35.29 & 59.10 & 62.95  & 46.10  \\
                    & LSTM~\citep{solomon_analyzing_2021}   & 30.26 (0.42) & 36.25 (0.18) & 60.86 (0.15) & 65.90 (0.27)  & 47.54 (0.13)  \\
                    & Deepmove~\citep{FengLZSMGJ18}         & 30.82 (0.19) & 36.27 (0.13) & 60.96 (0.13) & 66.06 (0.17) & 47.57 (0.10) \\
                    & MobTcast~\citep{xue2021mobtcast}      & \underline{32.29} (0.28) & 37.20 (0.29) & 60.34 (0.37) & 65.51 (0.35)  & 47.76 (0.30)  \\
                    & LSTM attn                             & 32.16 (0.48) & \underline{37.73} (0.08) & \underline{61.65} (0.21) & \underline{66.57} (0.37) & \underline{48.70} (0.12)  \\
                    & Ours                                  & \textbf{34.89} (0.16) & \textbf{39.93} (0.14) & \textbf{61.89} (0.17) & \textbf{66.92} (0.10) & \textbf{49.94} (0.11) \\ \midrule

\multirow{6}{*}{Yumuv} & 1-MMC~\citep{Gambs_2012}          & 34.81                    & 45.00 & 68.61 & 71.58  & 55.72  \\
                    & LSTM~\citep{solomon_analyzing_2021}   & 41.46 (0.34)             & 45.32 (0.26)             & 68.72 (0.23)             & 72.25 (0.29)             & 56.10 (0.22) \\
                    & Deepmove~\citep{FengLZSMGJ18}         & 42.29 (0.21)             & 46.11 (0.26)             & 69.47 (0.08)             & 73.01 (0.06)             & 56.91 (0.19) \\
                    & MobTcast~\citep{xue2021mobtcast}      & 42.65 (0.28)             & 46.13 (0.25)             & 69.18 (0.17)             & 72.81 (0.07)             & 56.69 (0.16) \\
                    & LSTM attn                             & \underline{43.02} (0.28) & \underline{46.66} (0.23) & \underline{69.87} (0.11) & \underline{73.48} (0.16) & \underline{57.38} (0.15) \\
                    & Ours                                  & \textbf{45.43} (0.32)    & \textbf{48.67} (0.20)    & \textbf{70.17} (0.09)    & \textbf{73.50} (0.08)    & \textbf{58.49} (0.13) \\ \bottomrule

\end{tabular}
\end{table*}

We implement several popular methods as baselines to compare with our proposed model. 
%
%
The approaches we choose are as follows: 
(1) \textbf{Markov models}. The \textit{de facto} standard method for this task~\citep{huang_mining_2017, Gambs_2012, KulkarniG19}. We implement the 1\textsuperscript{st} order Mobility Markov Chain (1-MMC)~\citep{Gambs_2012}, as our pre-experiment shows that higher-order methods can not improve the prediction performance. 
(2) \textbf{LSTM network}, a widely adopted recurrent neural network (RNN) model for sequence data, has been successfully applied to predict the next location~\citep{solomon_analyzing_2021, xu_understanding_2022}. We implement the model proposed by~\citet{solomon_analyzing_2021}. 
(3) \textbf{Deepmove}~\citep{FengLZSMGJ18}. It incorporates a historical attention module to extract historical movement patterns with RNN-based networks. We consider the most recent two days as the current sequence and the remaining five days as the historical sequence for the input to the model. 
(4) \textbf{MobTcast}~\citep{xue2021mobtcast}. It considers temporal, semantic, social, and geographical contexts with a transformer encoder-based structure to forecast the next POI. Also, the model includes two losses to encourage predicted and ground truth locations to be close in space. We incorporate all components except for the social context since the overlap between location visits for our users is low due to the high-resolution GPS tracking. 
(5) \textbf{LSTM with self-attention (LSTM attn)}. Inspired by~\citet{li_hierarchical_2020}, we implement an LSTM-based network with self-attentions between the current and previous hidden states.

We use the following metrics to evaluate the next location prediction performance of various methods: 
%
(1) \textbf{Accuracy}. It indicates the number of correctly predicted locations by the network. We rank the predicted probability from $P(\hat{l_{n+1}})$ in descending order and count the number of times that the ground truth location appears in the top-k predicted locations (Acc@k). Acc@1, Acc@5, and Acc@10 are reported. 
%
(2) \textbf{F1 score}. Since certain locations are visited more often than others~\citep{song_modelling_2010}, we employ the F1 score weighted by location visitation frequency. We argue that the F1 score is a better metric than Acc@1 as it considers the unbalanced location visits. 
%
(3) \textbf{Mean Reciprocal Rank (MRR)}. The metric is commonly applied to measure performance in information retrieval and re-identification tasks. MRR is the harmonic mean of the ground truth label's rank in the prediction.

\subsection{Results}
\subsubsection{Overall performance}

Table~\ref{tab:performance} shows the performance of next place prediction methods on the two considered datasets. We train each DL model 5 times with different random seeds and report the mean and the standard deviation of the respective performance indicators.

The 1-MMC method provides a strong baseline for the task. However, its performance is worse than any DL model, indicating that the Markov property cannot fully capture mobility patterns. 
On the contrary, the naïve LSTM model outperforms the 1-MMC method by a relatively large margin, demonstrating the effectiveness of DL-based models in predicting mobility. The worse performance compared to other DL methods can be attributed to the insufficient consideration of long-term dependencies in location visit sequences, an inherent shortcoming of the LSTM model~\citep{Vaswani_2017}. 
This deficiency is partly tackled by introducing the historical attention module in the Deepmove model that explicitly focuses on extracting historical information, which consistently outperforms the naïve LSTM on all indicators for both datasets. 
MobTcast obtains comparable performance in Yumuv and slightly better F1 and Acc@1 scores in GC compared to Deepmove, suggesting the additional consideration of context and the introduction of two losses have limited effect on predicting an individual's next location.

However, we observe a significant performance increase by adding a self-attention mechanism to the naïve LSTM model. The self-attention's ability to focus on specific steps during prediction indicates that successfully mining historical patterns can greatly benefit location prediction. 
By incorporating the multi-head self-attention and introducing the next mode prediction loss, the proposed model outperforms all other models by a relatively large margin. The relative increase of the F1 score is $8.05\%$ and $5.60\%$ for two datasets, respectively, which is calculated using the formula $(F1_1-F1_2)/F1_2$, where $F1_1$ is the F1 score of our model and $F1_2$ represents the F1 score of the second-best performing model. The consistent pattern for both considered datasets demonstrates the effectiveness of our network architecture and loss design.
Additionally, we report that the performance difference between LSTM with self-attention and our transformer decoder-based model is relatively small in Acc@5 and Acc@10 compared to the other indicators. We assume that the single head self-attention can already extract multiple likely visited next locations but lacks the ability to distinguish the importance within this set successfully.

\subsubsection{Ablation study}

\begin{table*}[htbp!]

\caption{Performance of the ablation study for next location prediction. We consider the model with (\checkmark) and without (-) adding temporal features (T), the travel mode feature (F), and the mode prediction loss ($\Lagr_2$).}
\label{tab:loc_ablation}
\begin{tabular}{@{}ccccccccc@{}}
\toprule
\multirow{2}{*}{Dataset} & \multicolumn{3}{c}{Module}                      & \multirow{2}{*}{F1} & \multirow{2}{*}{Acc@1} & \multirow{2}{*}{Acc@5} & \multirow{2}{*}{Acc@10} & \multirow{2}{*}{MRR} \\
                         & T                 & F            & $\Lagr_2$    &    \\ \midrule
\multirow{7}{*}{GC}      & -                 & -            & -            &  31.24 (0.18)             & 36.22 (0.22)             & 60.57 (0.25)             & 65.75 (0.23)             & 47.42 (0.24)             \\
                         & -                 & -            & \checkmark   &  31.20 (0.18)             & 36.64 (0.19)             & 60.99 (0.10)             & 66.09 (0.07)             & 47.83 (0.12)             \\
                         & -                 & \checkmark   & -            &  31.61 (0.23)             & 36.80 (0.18)             & 60.88 (0.07)             & 65.98 (0.10)             & 47.86 (0.09)             \\
                         & \checkmark        & -            & -            &  34.55 (0.19)             & 39.10 (0.20)             & 61.10 (0.24)             & 66.18 (0.19)             & 49.07 (0.17)               \\
                         & \checkmark        & -            & \checkmark   &  \underline{34.80} (0.29) & \underline{39.81} (0.12) & \underline{61.75} (0.11) & \underline{66.67} (0.10) & \underline{49.76} (0.07)  \\
                         & \checkmark        & \checkmark   & -            &  34.60 (0.09)             & 39.39 (0.10)             & 61.23 (0.07)             & 66.28 (0.16)             & 49.34 (0.08)   \\
                         & \checkmark        & \checkmark   & \checkmark   &  \textbf{34.89} (0.16)    & \textbf{39.93} (0.14)    & \textbf{61.89} (0.17)    & \textbf{66.92} (0.10)    & \textbf{49.94} (0.11)     \\\midrule
\multirow{7}{*}{Yumuv}   & -                 & -            & -            &  42.11 (0.17)             & 45.90 (0.12)             & 69.47 (0.12)             & 73.12 (0.12)             & 56.73 (0.09)             \\
                         & -                 & -            & \checkmark   &  42.83 (0.24)             & 46.45 (0.19)             & 69.47 (0.17)             & 72.90 (0.11)             & 57.01 (0.13)             \\
                         & -                 & \checkmark   & -            &  42.68 (0.18)             & 46.30 (0.18)             & 69.68 (0.14)             & 73.15 (0.18)             & 57.01 (0.16)             \\
                         & \checkmark        & -            & -            &  44.53 (0.10)             & 47.78 (0.05)             & 69.68 (0.27)             & 73.26 (0.19)             & 57.82 (0.10)             \\
                         & \checkmark        & -            & \checkmark   &  \underline{45.04} (0.22) & \underline{48.37} (0.19) & \underline{70.08} (0.12) & \underline{73.41} (0.14) & \underline{58.28} (0.12) \\
                         & \checkmark        & \checkmark   & -            &  44.65 (0.11)             & 47.97 (0.11)             & 69.86 (0.12)             & 73.38 (0.12)             & 58.01 (0.10)             \\
                         & \checkmark        & \checkmark   & \checkmark   &  \textbf{45.43} (0.32)    & \textbf{48.67} (0.20)    & \textbf{70.17} (0.09)    & \textbf{73.50} (0.08)    & \textbf{58.49} (0.13)    \\ \bottomrule
\end{tabular}
\end{table*}

We perform an extensive ablation study to understand the importance of each component in our proposed model. In Table~\ref{tab:loc_ablation}, we show the performance of different variants of our model, considering whether or not to include temporal features $h_k$ and $d_k$, the travel mode feature $e_k$, and the next travel mode prediction loss $\Lagr_2$. The performance results are consistent across both datasets. 
We first report that the model performs significantly better with temporal features. This increase is intuitive as the time encodes different levels of periodicity in location visits compared to merely having the sequence information. While considering the travel mode feature and the mode prediction loss are both beneficial for the model's performance, the latter is more important for the task. 
Moreover, we observe a further performance gain when simultaneously including the travel mode feature and the mode prediction loss, as indicated by the highest performance achieved by the complete model for both datasets. 
These experiments show that temporal features are indispensable for accurate next location prediction. 
They also confirm that both historical travel mode patterns and the ability to predict the next travel mode are essential for the task, justifying the effectiveness of our model design.

Since our proposed model outputs both the probability of the next location and the next travel mode, we can utilize the same model structure to evaluate the performance of predicting the next travel mode. From this viewpoint, time and location can be regarded as additional features, and predicting the next location can be seen as the auxiliary task. We evaluate the usefulness of these components in the next mode prediction task using an ablation study. The performance result are shown in Table~\ref{tab:mode_ablation}. We again observe a large performance increase after including temporal features. Furthermore, historical location visits and the additional location prediction loss benefit the prediction. The top performance is achieved by the complete model that utilizes all three components in the network.
These results indicate that the next travel mode prediction performance can be improved by considering location visit information. Combined with the location prediction ablation study, we conclude that the next location and the next travel mode prediction tasks are inherently dependent and should be tackled together to achieve optimal performance.

\begin{table}[htbp!]
\tabcolsep=0.13cm

\caption{Performance for next mode prediction. We consider the model with and without adding temporal features (T), the location feature (F), and the location prediction loss ($\Lagr_1$).}
\label{tab:mode_ablation}
\begin{tabular}{@{}cccccccc@{}}
\toprule
\multirow{2}{*}{Dataset} & \multicolumn{3}{c}{Module}                       & \multirow{2}{*}{F1} & \multirow{2}{*}{Acc@1} & \multirow{2}{*}{MRR}    \\
                         & T                 & F            & $\Lagr_1$    &    \\ \midrule
\multirow{4}{*}{GC}      & -                 & -            & -            & 54.29 (0.16)             & 58.17 (0.09)             & 75.95 (0.05)              \\
                         & \checkmark        & -            & -            & 57.30 (0.18)             & 60.14 (0.09)             & 77.28 (0.04)       \\
                         & \checkmark        & \checkmark   & -            & \underline{58.64} (0.33) & \underline{61.03} (0.29) & \underline{77.67} (0.17) \\
                         & \checkmark        & \checkmark   & \checkmark   & \textbf{58.75}    (0.08) & \textbf{61.16} (0.25)    & \textbf{77.72} (0.16)  \\\midrule
\multirow{4}{*}{Yumuv}   & -                 & -            & -            & 50.84 (0.12)             &  52.53 (0.13)             & 71.90 (0.07)              \\
                         & \checkmark        & -            & -            & 52.59 (0.11)             &  53.96 (0.13)             & 72.74 (0.09)              \\
                         & \checkmark        & \checkmark   & -            & \underline{53.70} (0.21) &  \underline{54.64} (0.10) & \textbf{73.05} (0.06)  \\
                         & \checkmark        & \checkmark   & \checkmark   & \textbf{54.03} (0.19)    &  \textbf{54.84} (0.11)    & \textbf{73.05} (0.08) \\ \bottomrule
\end{tabular}
\end{table}

\subsubsection{Importance of travel mode information}

\begin{figure}[htbp!]
  \centering
    \begin{subfigure}[]{0.49\linewidth}
    \includegraphics[width=\linewidth]{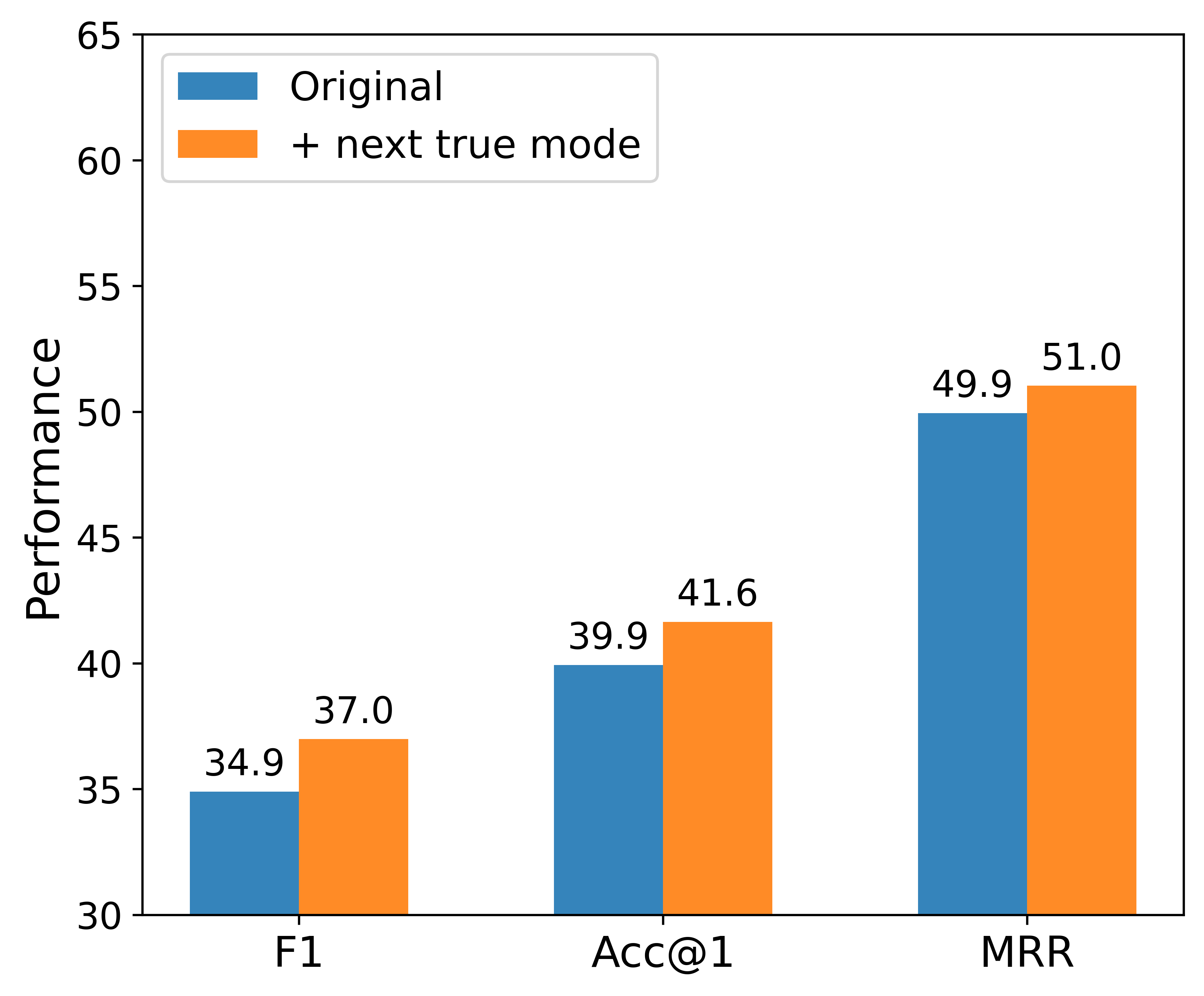}
    \caption{Green Class}
    \end{subfigure}
    \hfill
    \begin{subfigure}[]{0.49\linewidth}
    \includegraphics[width=\linewidth]{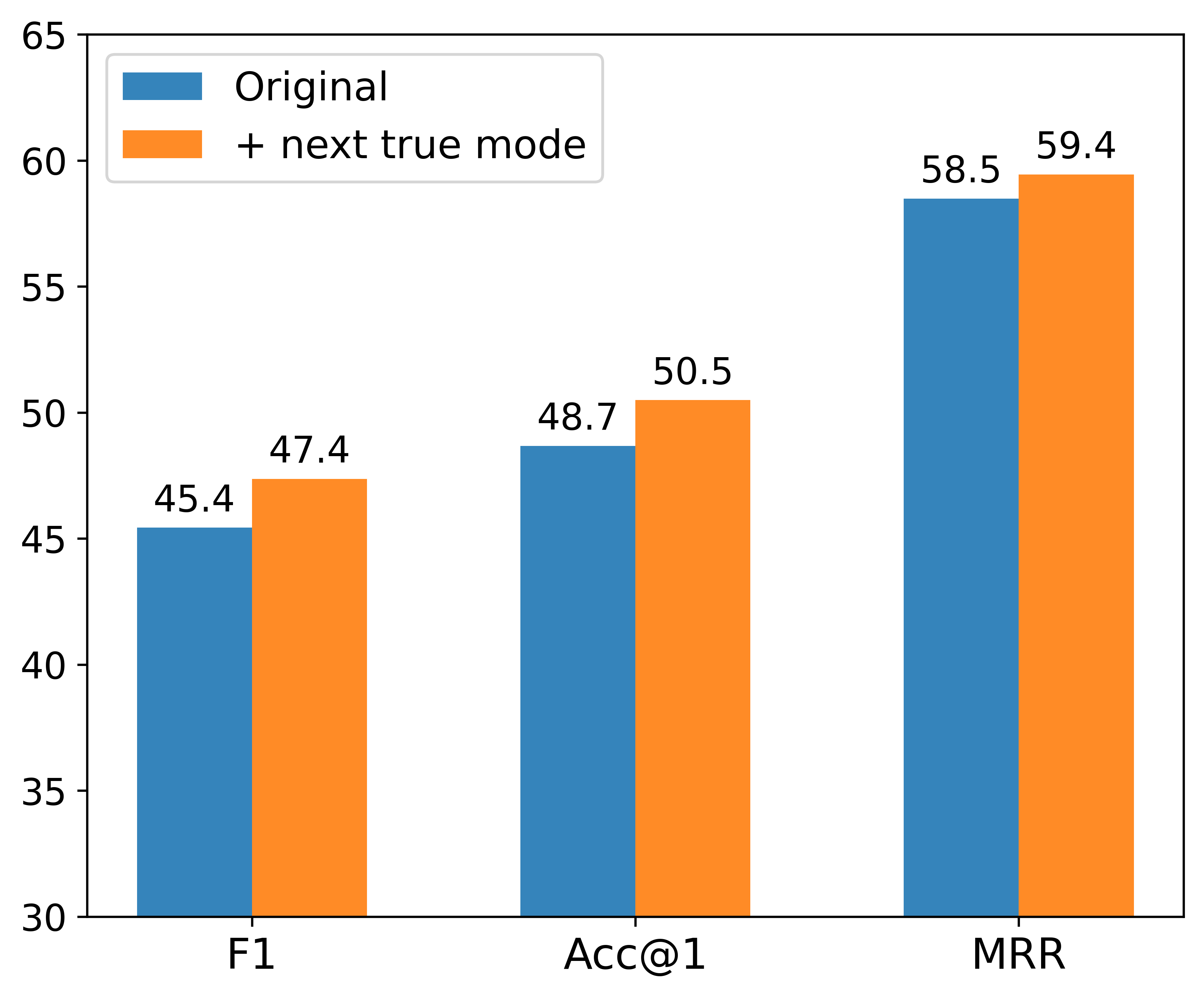}
    \caption{Yumuv}
    \end{subfigure}%

  \caption{Performance of the original model (blue) and the model that considers the ground truth next travel mode (orange).}
  \label{fig:next_mode}
\end{figure}

We further investigate the influence of travel mode on the next location prediction task. 
To understand to what extent the next location prediction result depends on the prediction of travel mode, we input the ground truth next travel mode into the model. 
This is achieved by feeding the travel mode at the next time step into the mode embedding layer to generate the embedding vector $emb^{e}_{n+1}$. Then, the vector is added to the user embedding and further processed with the fully connected layer, i.e., Eq.~(\ref{equation:fc}) becomes $f_n = FC((out_{n} + \text{Linear}(emb^{e}_{n+1})) \oplus emb^{u^{(i)}})$. 
We can regard this new problem as predicting the next location conditioned on the next travel mode, where the scenario is to predict the location when knowing which travel mode the user will take. 
Figure~\ref{fig:next_mode} compares the performance of the original model with the model that includes the next travel mode. When knowing the next travel mode, we observe a notable increase in all indicators for the two datasets. In line with the conclusion from the ablation study, this result illustrates that the choice of the next location depends strongly on the travel mode. 
The performances shown in Figure~\ref{fig:next_mode} can also be interpreted as the upper bound of performance gain that can be achieved by including the prediction of travel mode in our model. This result suggests that working on a better mode prediction still holds great potential to increase the prediction accuracy of the next location.

\begin{figure}[htbp!]
  \centering
    \begin{subfigure}[]{0.49\linewidth}
    \includegraphics[width=\linewidth]{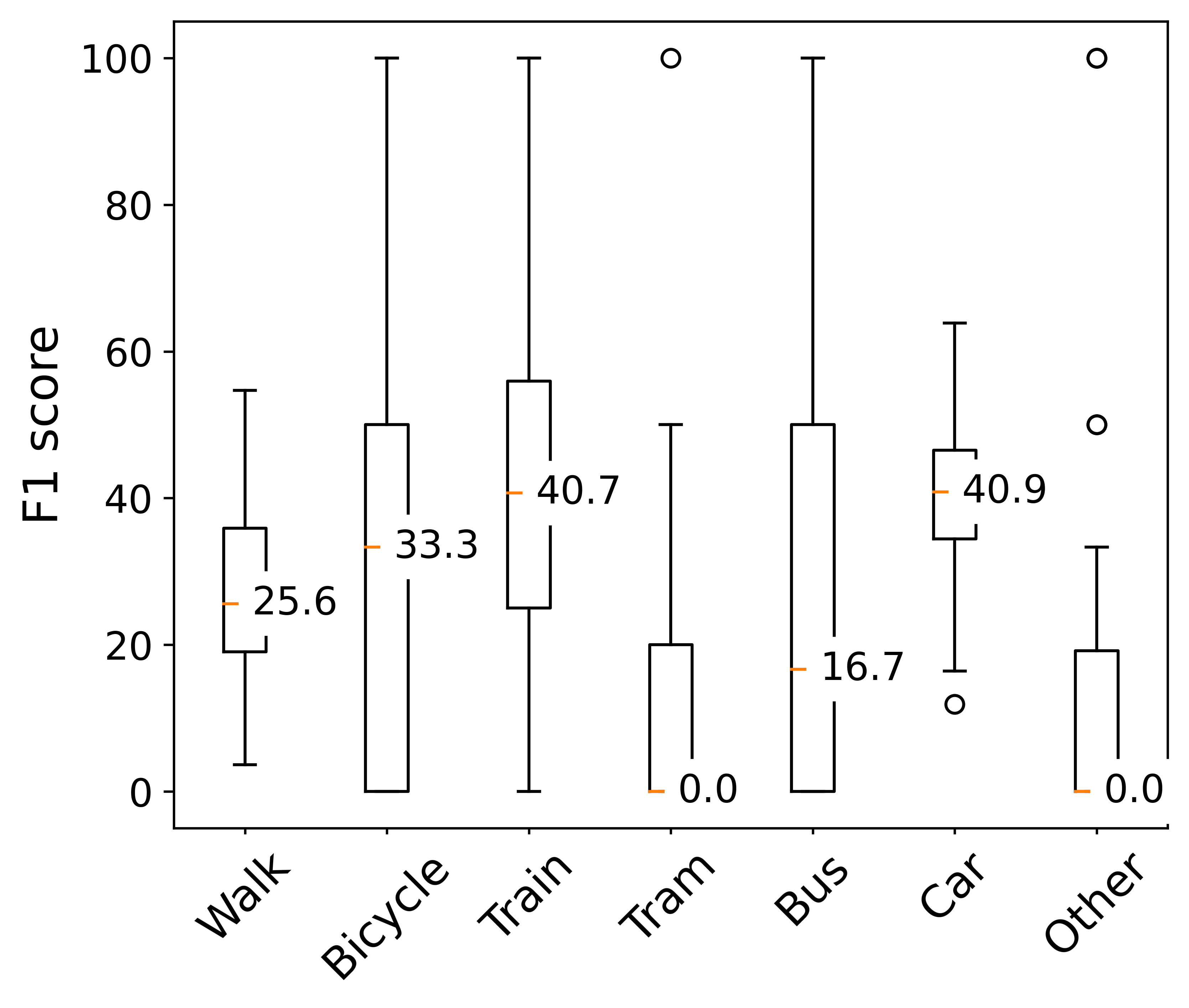}
    \caption{GC}
    \end{subfigure}
    \hfill
    \begin{subfigure}[]{0.49\linewidth}
    \includegraphics[width=\linewidth]{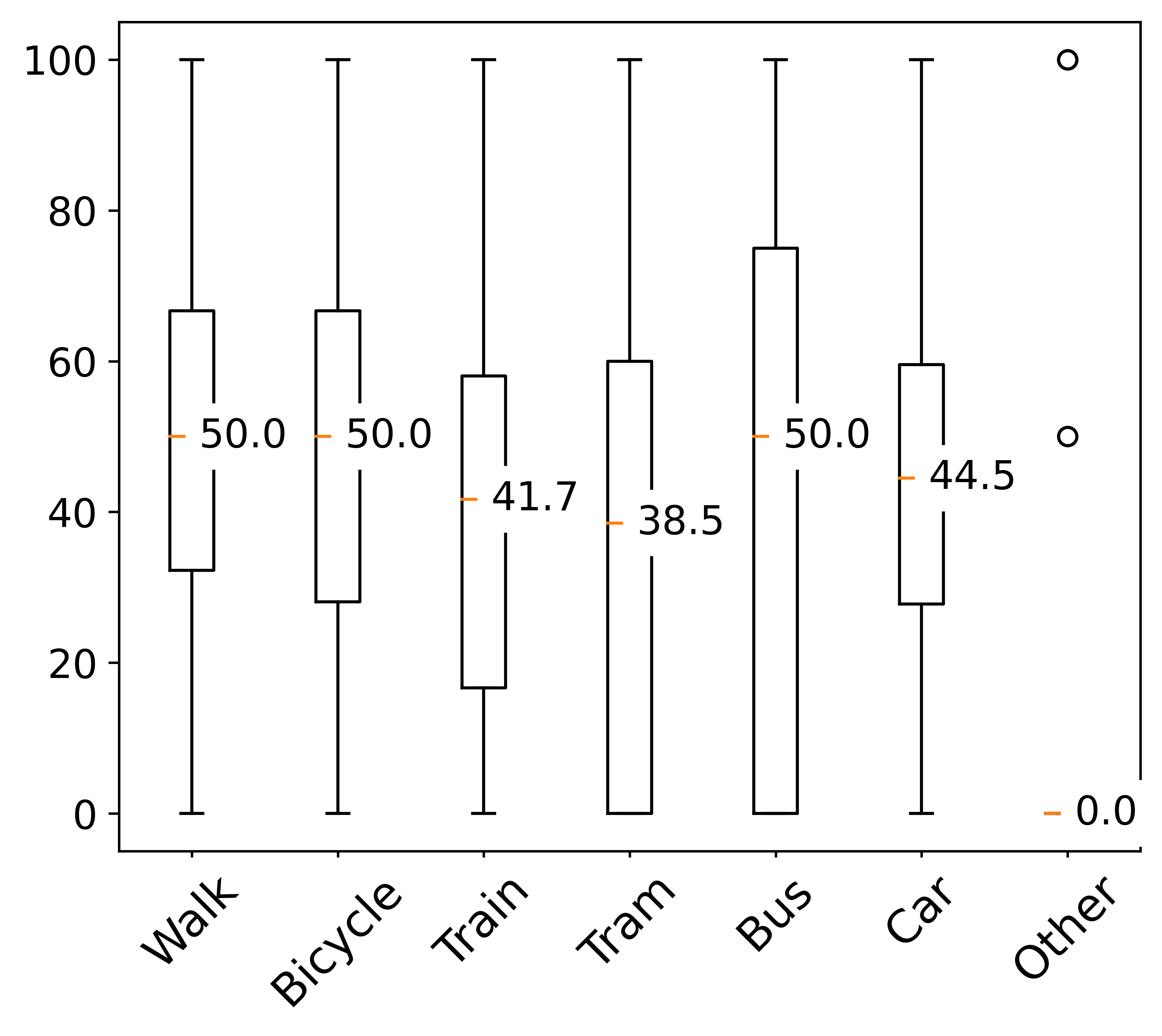}
    \caption{Yumuv}
    \end{subfigure}%
  \caption{F1 score boxplot of individual categorized by the next travel mode.}
  \label{fig:F1_mode}
\end{figure}

This mode-location dependency is evident by checking differences in predicting various categories of locations. 
In particular, we group the location prediction result based on the ground truth next travel mode and plot the F1 score in Figure~\ref{fig:F1_mode}. Each ``box'' in the figure is generated by the F1 score of every user and represents the model's performance in a single travel mode. We also labelled each box with the median F1 score across users. 
The differences in the F1 scores suggest that location prediction depends on the travel mode: locations with specific travel modes are more challenging to predict than others. For example, the locations reached by \textit{tram} and \textit{bus} obtained relatively lower performance than other modes in GC (Figure~\ref{fig:F1_mode}a). For Yumuv, the mode \textit{tram} achieves a relatively low median F1-Score, and the mode \textit{bus} has a high variance across users (Figure~\ref{fig:F1_mode}b). 
This observation could be attributed to the more variate choice of locations when users travel with these modes. The failure to predict the mode \textit{other} (including \textit{airplane}, \textit{ski} and \textit{coach}) is related to the exploration nature of human mobility: users often travel with these modes to new locations that are difficult for the model to predict. Moreover, the distinct pattern for the GC and Yumuv datasets can be explained by the mobility patterns of the two groups of participants. GC participants are more active in their mobility and often travel a longer distance by car and train~\citep{Hong_2022}, whereas Yumuv users mostly commute within the city~\citep{reck_mode_2022} and have a more balanced use of travel modes (Table~\ref{tab:number}).

\section{Discussion and Conclusion}
Travel behaviour researchers have long realized that individuals' choice of locations is highly influenced by other travel behaviour dimensions, such as the day of the week and the availability of travel modes. While this dependency is evident in an increasing number of empirical studies, limited attention has been placed on integrating this information into the next location prediction task. 
In this study, we present a transformer-based model to predict the next location visit of an individual. In particular, the model is designed to accurately predict the next location and the next travel mode at the same time. 
We test our model on two large-scale real-world GPS tracking datasets and compare its performance with most recent methods for the task. Our experiments show that the proposed model learns the dynamics of human mobility from historical time, travel mode and location sequences and obtained state-of-the-art next location prediction performance. 
Our extensive ablation study demonstrates that temporal and travel mode features and the ability to predict the next travel mode are essential components in an accurate location prediction model. 
Moreover, our analysis indicates strong mode-location dependency, which suggests the next location and the next travel mode prediction tasks should be tackled together to achieve optimal performance.

We conclude that considering historical travel mode patterns increases the accuracy of the next location prediction. This result connects travel mode detection models with next location prediction models. Mode detection models aim to classify a movement's travel mode given its spatio-temporal characteristics~\citep{roy_assessing_2022}, which can enrich general tracking datasets that lack accurate user-provided travel mode labels. Therefore, mobility prediction can be tackled using a two-step approach: (1) identify the travel mode with detection models based on spatio-temporal movement characteristics; and (2) predict the next location with prediction models based on temporal, travel mode and location information. This approach indicates that an optimal system for mobility prediction should develop accurate models for both tasks. 

Next location prediction is an essential backbone of many sustainable transport solutions; however, it is a challenging problem that is not yet fully tackled. Although we improve the state-of-the-art considerably compared to the other methods, the field still needs more research to achieve a more accurate prediction performance. We propose several future directions based on the results of this work. 
This study demonstrates the importance of travel mode for predicting an individual's immediate next location visit. Since the mode-location dependency is more evident on a longer time scale~\citep{susilo_repetitions_2014, Hong_2022}, future research should consider integrating travel mode information into mobility simulation models for generating more realistic mobility sequences. 
Furthermore, besides the interdependent travel behaviour dimensions, studies have shown that external contexts, such as the built environment, play an indispensable role in shaping individuals' mobility~\citep{farinloye_qualitatively_2019}. This knowledge can be effectively extracted and combined with individual mobility in DL models based on land-use maps and POI data. Whether or not the additional information will improve the performance of next location prediction or mobility generation tasks is a direction worth exploring.

\begin{acks}
    This work is supported by the Hasler Foundation (grant number 1-008062). The yumuv dataset was collected as part of a joint study with Swiss Federal Railways that was financed through the ETH Mobility Initiative (grant number MI-01-19).
\end{acks}

\bibliographystyle{ACM-Reference-Format}
\bibliography{base}



\end{document}